\documentclass[sigconf,authorversion=true]{aamas}  

\usepackage{booktabs}
\usepackage{algorithm}
\usepackage[noend]{algpseudocode}
\usepackage{amsmath,amscd}
\usepackage{etoolbox}

\setcopyright{ifaamas}  
\acmDOI{}  
\acmISBN{}  
\acmConference[AAMAS'18]{Proc.\@ of the 17th International Conference on Autonomous Agents and Multiagent Systems (AAMAS 2018)}{July 10--15, 2018}{Stockholm, Sweden}{M.~Dastani, G.~Sukthankar, E.~Andr\'{e}, S.~Koenig (eds.)}  
\acmYear{2018}  
\copyrightyear{2018}  
\acmPrice{}  

\begin{document}

\title{One-Shot Learning using Mixture of Variational Autoencoders:\\a Generalization Learning approach}  



\author{Decebal Constantin Mocanu}
\affiliation{%
  \institution{Eindhoven University of Technology}
  \city{Eindhoven} 
  \state{Netherlands} 
  \postcode{5612 AP}
}
\email{d.c.mocanu@tue.nl}
\author{Elena Mocanu}
\affiliation{%
  \institution{Eindhoven University of Technology}
  \city{Eindhoven} 
  \state{Netherlands} 
  \postcode{5612 AP}
}
\email{e.mocanu@tue.nl}

\begin{abstract}  
Deep learning, even if it is very successful nowadays, traditionally needs very large amounts of labeled data to perform excellent on the classification task. In an attempt to solve this problem, the one-shot learning paradigm, which makes use of just one labeled sample per class and prior knowledge, becomes increasingly important. In this paper, we propose a new one-shot learning method, dubbed MoVAE (Mixture of Variational AutoEncoders), to perform classification. Complementary to prior studies, MoVAE represents a shift of paradigm in comparison with the usual one-shot learning methods, as it does not use any prior knowledge. Instead, it starts from zero knowledge and one labeled sample per class. Afterward, by using unlabeled data and the generalization learning concept (in a way, more as humans do), it is capable to gradually improve by itself its performance. Even more, if there are no  unlabeled data available MoVAE can still perform well in one-shot learning classification. We demonstrate empirically the efficiency of our proposed approach on three datasets, i.e. the handwritten digits (MNIST), fashion products (Fashion-MNIST), and handwritten characters (Omniglot), showing that MoVAE outperforms state-of-the-art one-shot learning algorithms.
\end{abstract}

%


\maketitle


\section{Introduction}
Object recognition is an important problem, and it has many applications, e.g. computer vision \cite{Dalal2005,7410370,7298854,thewlis17Bunsupervised}, robotics~\cite{LONCOMILLA2016499} and healthcare~\cite{corr/LitjensKBSCGLGS17}. Traditional solutions use classifiers built on large amounts of data. In a time with more and more unlabeled data, manually labeling of all these data is costly, time consuming, and inefficient. Hence, the one-shot learning paradigm becomes increasingly important. The aims of this paradigm is to improve the generalization capabilities of the learning models (or algorithms) in such a way that they are capable to achieve a very good performance by having just one labeled sample per class or (at maximum) few labeled samples. To achieve this, usually the state-of-the-art one-shot learning algorithms make use of prior knowledge and large amounts of unlabeled data. Even if serious progress have been made in the last period on this paradigm, still its algorithms are usually immature and in an incipient phase. Thus,  there is place for many improvements. For instance, MNIST digits dataset may be considered to be overused and very simple by the scientific community nowadays. Yet, up to our best knowledge, the maximum classification accuracy achieved on MNIST by one-shot learning algorithms with one labeled sample per class (1-shot) is just about 72\%.

In this paper, we address the above problem, and we propose a new one-shot learning classification method, dubbed Mixture of Variational Autoencoders (MoVAE). Contrary to the state-of-the-art one-shot learning methods, MoVAE does not need at all any prior knowledge. In fact, it complements these methods. It starts from zero knowledge and one (or few) labeled samples per class, and then it gradually learns to generalize its knowledge using the generalization learning concept~\cite{WATERMAN1970}. Also, by opposite to the usual direction in artificial neural networks, MoVAE is not an unitary neural network. In fact, it is composed by many Variational Autoencoders (VAEs), each one learning the distribution of a class. Thus, MoVAE can be a good example of collective intelligence. Each VAE took separately can not perform classification, but all of them acting together, are able to learn and classify objects very well.

To assess the performance of our proposed method, we performed empirical studies on three different types of object recognition problems, i.e. digits recognition (MNIST dataset), fashion products classification (Fashion-MNIST dataset~\cite{xiao2017FashionMNIST}), and handwritten characters recognition (Omniglot dataset~\cite{salutomniglot}). Aiming to rise the one-shot learning performance, we analyzed the MoVAE behavior in different settings. To the end, we proved its advantages in comparison with state-of-the-art.

The remainder of this paper is organized as follows. Section~2 briefly presents some background information on one-shot learning and Variational Autoencoders~\cite{Kingma2013} for the benefit of the non-specialist reader. Section~3 introduces our proposed methods. Section~4 details the experiments performed and the results obtained, while Section~5 concludes the paper.

\section{Background}
\subsection{One-Shot learning}
In a time when we have access to big unlabeled datasets, learning from one or just few examples is essential. By the opposite, the most advanced machine learning methods use large labeled datasets in order to learn to classify useful object representations. However, in the last decade, starting with the work of \cite{FeiFei2006}, there are many approaches which propose solutions for one-shot learning, e.g. \cite{MillerCVPR2000,LakeScience2015,Koch2015SiameseNN,HariharanG16,VinyalsBLKW16}. 
As deep learning became more successfully, in the last few years, a bridge between one-shot learning and deep learning has been developed. The Siamese Net~\cite{Koch2015SiameseNN} was used to learn useful representations and a distance metric to say whether a test image and a given image belong to the same class or not.
Further on, Santoro et al. proposed a memory-augmented neural network in~\cite{SantoroBBWL16} to rapidly assimilate new data and by leveraging  this  data  to  make  accurate  predictions after seeing only just few  labeled samples. Maybe one of the most advanced one-shot learning models, i.e Matching Nets, has been proposed in~\cite{VinyalsBLKW16} and employs ideas from both, neural networks with external memories and metric learning. Even though there are many one-shot learning models now, it is interesting to mention here also a new direction which combines reinforcement learning with one-shot learning~\cite{WoodwardF17}.

For clarity, we mention that independently of the used model and dataset, there are two commonly utilized terms in one-shot learning. We will define them here and use them further throughout the paper. These terms are: (1) \textit{N-way} classification, which means that N classes from the dataset are considered to perform one-shot learning. Usually N is smaller than the total number of classes in the dataset, as the remaining ones are used to build prior knowledge; and (2) \textit{k-shot} learning which means that from each of the N classes, just k labeled samples are used for one-shot learning. If $k=1$ then one-shot learning is performed, and if $k>1$ then, in fact,  few-shot learning is performed.

\subsection{Variational auto-encoders (VAE)}
Based upon the auto-encodes procedure, variational auto-encoders \cite{Kingma2013} are providing to be powerful generative models used to reconstruct an output from an input. Their generalization capabilities arise from the adding of a sampling layer between encoder and decoder. Many variations of VAE appeared latter on, such as  hierarchical nonparametric variational autoencoders, which combines Bayesian nonparametric priors with VAEs \cite{corr/Goyal17}.

Let us consider $X=\{x^{(i)}\}_{i=1}^N$ our observable variable, and $z$ a latent variable (continuous). Thus a typical VAE model follows the next steps. Initially, a latent variable model is constructed in order to learn a mapping from some latent variable to an input distribution, such that $x \xrightarrow[]{f(z)}z$   by training a neural network, where $f(x)$ is a  joint probability density function, $p_{\theta}$, over the network parameters, $\theta$, such that $f(z) =p_{\theta}(x|z)$. Thus,
\begin{equation}
p(x)=\int p(x,z) dz 
\end{equation}
where $p(x,z)=p(x|z)p(z)$. 

\subsubsection{Variational inference}

Having a relatively similar idea with the Helmholtz machine~\cite{helmholtzDayan95} or wake-sleep algorithm~\cite{wakesleepHinton1995}, in VAE models \cite{Kingma2013,pmlr-v32-rezende14} the true posterior $q_{\phi}(z|x)$ is aproximated using variational parameters, $\phi$. Given the intractable posterior $p_{\theta}(z|x)$, the VAE introduce an inference model $q_{\phi}(z|x)$ (parametrized with another neural network) that learns to approximate the posterior by optimizing the variational lower bound, such that $p_{\theta}\ge \mathcal{L}(\theta,\phi,x)$. Hence, the objective function is a sum over the reconstruction error and the regularization terms, given by
\begin{equation}\label{eq:loss} 
\mathcal{L}=-D_{KL} (q_{\phi}(z|x)||p_{\theta}(z))+E_{q_{\phi}(z|x)}[\log p_\theta(x|z)] 
\end{equation}
The first term in equation~\ref{eq:loss} is the Kullback-Leibler divergence between the inference model, $q_{\phi}(z|x)$, and the posterior distribution~\cite{Kullback59}. Further on we can consider the variational lover bound as $\mathcal{L}(\theta,\phi,x) \leq E_{q_{\phi}(z|x)}[\log p_\theta(x|z)]$ since the Kullback-Leibler divergence is always positive.
\subsubsection{Reparametrization trick}
 There are so far two types of connections, the top-down connections $\theta$ implementing the generative model and the bottom-up connections $\phi$ implementing the inference model. Finally a  reparametrization trick is used in the VAE models by adding a noise signal, $\epsilon$, into the latent variable.

\begin{equation*}
\begin{CD}
x @>{q_{\phi}(z|x)}>> z@<{p_{\theta}(z|x)}<< \hat{x}\\
@A{x_{aug}^*}AA @AA{\epsilon\sim p(\epsilon)}A@.\\
@. @.\\
\end{CD}
\end{equation*}
VAE simultaneously train both the generative model $p_{\theta}(x|z)$ and the inference model $q_{\phi}(z|x)$  by optimizing the variational bound using gradient backpropagation. Horewer, a complementary approach to variational inference is the Markov Chain Monte Carlo (MCMC) method, as detailed for example in \cite{icml2015_salimans15}. 

\section{Mixture of Variational AutoEncoders}
In this section, we present the details of our proposed method, dubbed Mixture of Variational AutoEncoders (MoVAE). We start by giving its intuition. Then we continue by describing its technical details, and its inference procedure.
\subsection{Intuition}
The intuition behind MoVAE is simple and it is inspired by human learning processes. People, when they learn new concepts, they do not manage too well to deal with large amounts of labeled data, but they are often extremely efficient to generalize across various conditions just from one example. Sometimes, they make use of prior acquired knowledge, and sometimes not. They start just from one example and gradually add new representations of that example (or situation) to its default category using generalization~\cite{Generalizationhumans}. At a different scale, the learning concept evolved through the human world into a collective intelligence behavior. The advances of human society 
were mainly made, not by super-humans, but by many humans, connected between them in a social network, sharing a set of values, and working together for a common goal. Moreover, a human is far to be one of the strongest animal in the world. In fact, it is quite weak, but humans collaborative way of being and personal specialization made from the human race one of the most successful in the word~\cite{humankind}.

Keeping the proportion, by analogy, we argue that in machine learning, we should not search for the most powerful model possible, but to create many specialized models, each being capable of doing well its specialized task. Then, these models working together will be able to fulfill a common goal, inaccessible for a singular model. In a way, in artificial intelligence, this approach is followed by ensembles and swarm intelligence, with the difference that each particle or ensemble could do a better or worse job on the common task, while in what we propose next, one singular model would achieve nothing.

These being said, and knowing that a Variational Autoencoder can represent very well a data distribution, in this paper, we propose to build a Mixture of Variational Autoenconders (MoVAE) to perform classification. In the specific case of classification, each VAE of the mixture will be very specialized and will learn the distribution of just one class by being trained on samples belonging just to its specific class. Thus, after the learning phase, our assumption is that each VAE will reconstruct very well unseen images belonging to its encoded class, but if images belonging to other classes will be reconstructed through it, then their reconstructed version will be not so good. And here come the trick of cooperative inference. Each VAE model is not able to discriminate if a given image belongs to its encoded class if it looks just of that image reconstructed version, but the mixture of VAEs it is. If we pass the same image through all the VAEs belonging to the mixture then we obtained a reconstructed version of the original image for any VAE. Then the class of the original image is given by the VAE which obtains the best reconstruction of the original image. Moreover, our assumption is that our proposed approach does not need many labeled images to learn well the class distributions. In fact, it can use just one labeled sample per class to encode in a decent manner the corresponding class in each VAE. Then, by using generalization learning and considering unlabeled data it will be able to gradually increase the quality of the encoded distributions, being capable to improve by itself its discriminative capabilities, as described next.

\subsection{Algorithm}
MoVAE method is briefly described in Algorithm~\ref{alg:dbntrain}. It starts by initializing its two specific meta-parameters ($\psi$ which represents the number of unlabeled data samples considered at each generalization iteration and the classes $C$ of the dataset) and the other meta-parameters characteristic to any VAE model (i.e. Algorithm~\ref{alg:dbntrain}, Line 1). Then, it creates a VAE model for each class $C_i$, which is trained on a small amount of randomly selected labeled samples (i.e. Algorithm~\ref{alg:dbntrain}, Lines 3-6). Please note, if we would like MoVAE to perform only pure supervised learning, we have just to consider all labeled samples belonging to the training set in this initialization phase and then we can stop the procedure here. 

In the case of one-shot learning, after this initialization phase, we go further to the generalization phase in which a number of recursive generalization iterations is performed (i.e. Algorithm~\ref{alg:dbntrain}, Lines 8-21). At each generalization iteration, MoVAE, firstly, reconstructs all the unconsidered unlabeled samples, computes the distance from these reconstructions to the original samples, and sorts the samples according with this distance (i.e. Algorithm~\ref{alg:dbntrain}, Lines 9-13). Secondly (i.e. Algorithm~\ref{alg:dbntrain}, Lines 14-19), it adds to the training set $x_i$ of each VAE$_{C_i}$ (the VAE model corresponding to class $C_i$) a number of $\psi/|C|$ samples (where $|C|$ is the total number of classes) which are best reconstructed by VAE$_{C_i}$ (i.e. Algorithm~\ref{alg:dbntrain}, Line 16). At the same time, these $\psi/|C|$ samples are badly reconstructed by the others VAEs (i.e. Algorithm~\ref{alg:dbntrain}, Line 17). This is done with the aim of avoiding introducing a propagating error through the generalization iterations, as much as possible. Each of these selected samples are then removed from the set of unconsidered unlabeled samples. Thirdly (i.e. Algorithm~\ref{alg:dbntrain}, Lines 20-21), each VAE$_{C_i}$ belonging to MoVAE is retrained using the new obtained $x_i$.

\begin{algorithm}
\caption{MoVAE schematic algorithm.}
\begin{algorithmic}[1]
\State Initialize meta-parameters, e.g. $\psi$,$C$
\State \%\%\textit{initialize all VAEs belonging to MoVAE}
\For {each class $C_i$ of the data set}
\State $x_i\gets$ 1 (or few) random samples
\State $creates$ VAE$_{C_i}\in$ MoVAE
\State train VAE$_{C_i}$ with $x_i$
\EndFor
\State \%\%\textit{generalization iterations}
\For {a specific amount of generalization iterations}
\For {each VAE$_{C_i}\in$ MoVAE}
  \For {all $u_k$ unconsidered unlabeled samples}
  \State $\widehat{u}_{k,i}\gets$ $u_k$ $reconstructed$ by VAE$_{C_i}$ 
  \State $compute$ the distance $d(\widehat{u}_{k,i},u_k)$
  \EndFor
  \State $sort$ $\widehat{u}_{k,i}$ and $u_k$ according with $d(\widehat{u}_{k,i},u_k)$
\EndFor
\For {each VAE$_{C_i}\in$ MoVAE}
\For {a number of $\psi/|C|$ samples}
\State $\eta\gets$ $u_k$ with $\min_k(d(\widehat{u}_{k,i},u_k)$)
\If {$\eta\notin$ $\widehat{u}_{0:\psi,C\setminus i}$}
\State $x_i\gets x_i\cup\eta$
\State $remove$ $\eta$ from $u_k$, $\widehat{u}_{k,i}$, and $d(\widehat{u}_{k,i},u_k)$
\EndIf
\EndFor
\EndFor
\For {each VAE$_{C_i}\in$ MoVAE}
\State $retrain$ VAE$_{C_i}$ with $x_i$
\EndFor
\EndFor
\end{algorithmic}
\label{alg:dbntrain}
\end{algorithm}

\subsection{Inference}
After the MoVAE model is trained, for inference, a very simple metric can be used. The basic idea is that the estimated class for a specific image is given by that VAE$_{C_i}$ which is capable to obtain the best reconstructed image. Thus, for a given image $\mu$ the following formula can be used:
\begin{equation}
 \mu_{class}\gets argmin_{i} d_{VAE_{C_i}}(\widehat{\mu}_{i},\mu)
\end{equation}
\begin{figure*}[t!]  
\centering
\includegraphics[width=0.8\columnwidth]{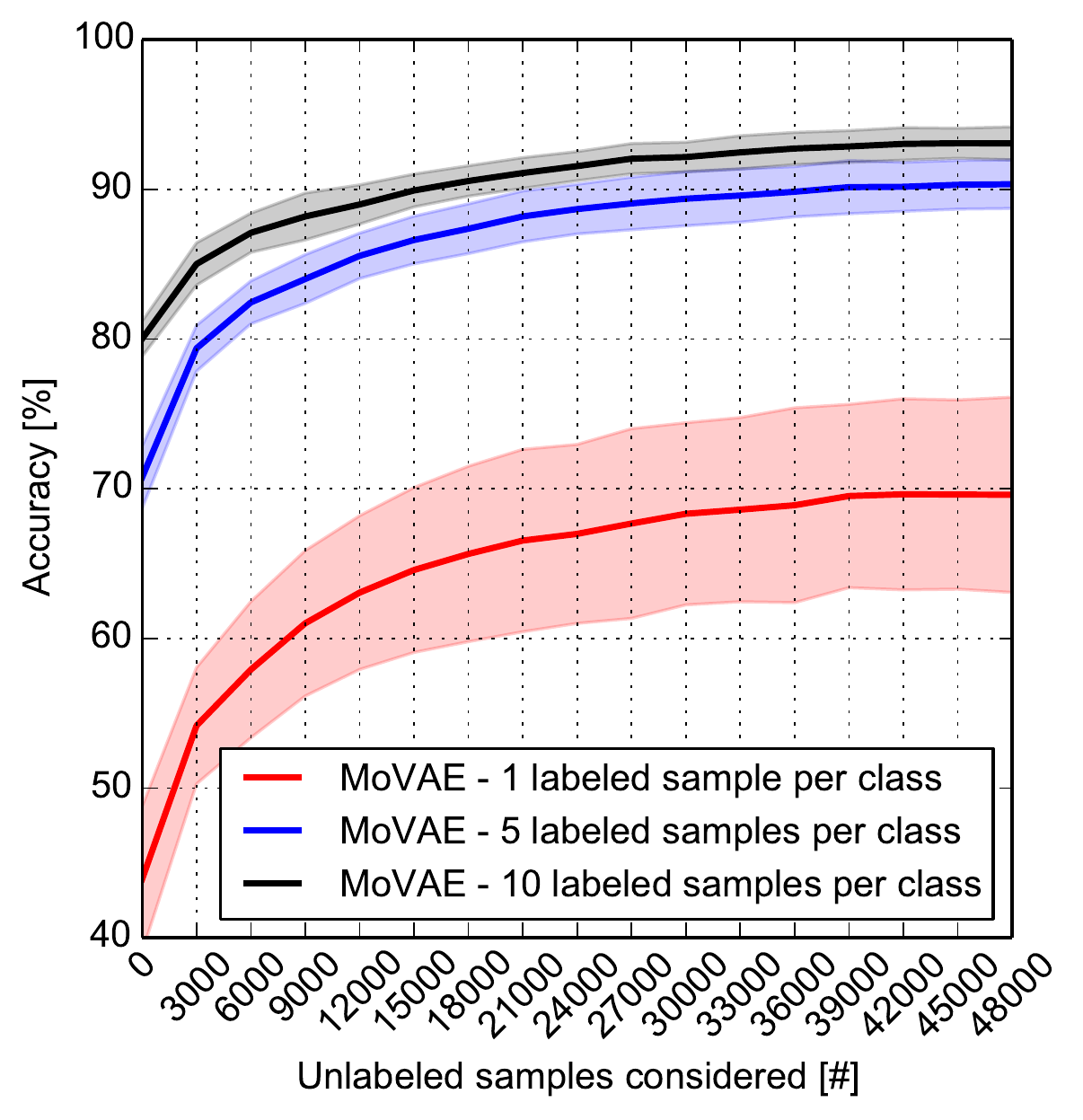}
\includegraphics[width=0.8\columnwidth]{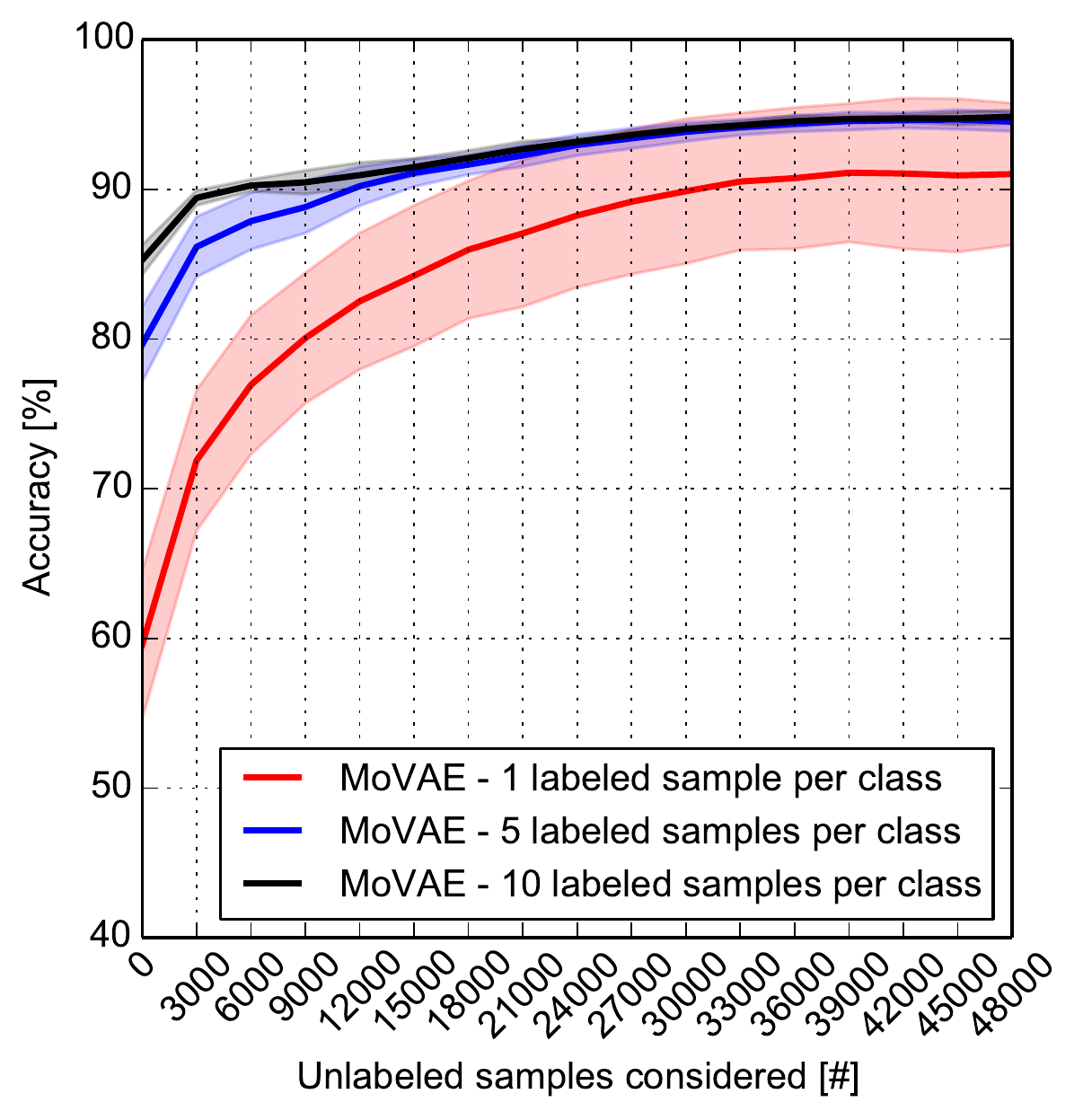}
\caption{One-shot learning. MoVAE performance on the test set of the MNIST dataset during training without data augmentation (left) and with data augmentation (right). At each generalization step 3000 previously unseen unlabeled samples were taken into consideration. The straight lines represent the mean values, while the shadowed areas represent the standard deviation.}
\label{fig:mnistoneshot}
\end{figure*}

To assess the distance between the image $\mu$ and its reconstructed version $d(\hat{\mu},\mu)$, more metrics are suitable, but we found out that Pearson Correlation Coefficient is a very good one, being fast and accurate. Moreover, it is good mentioning that during all phases (especially in the initialization phase), techniques such as data augmentation may be used to increase artificially the number of unlabeled or labeled samples, as we show further in the Experimental section.

\section{Experiments and results}

We assessed the performance of the proposed method on three datasets, proceeding in a step wise fashion. Firstly, we have evaluated MoVAE for pure supervised learning tasks (all available labeled training data were used) in Section 4.1. Secondly,  in Section 4.2 we performed experiments in an one-shot semi-supervised learning context. Thirdly, the final evaluation of MoVAE has been done in a pure one-shot learning fashion in Section 4.3.

\subsection{Purely supervised learning}

As MoVAE is also a new classification model, first we have assessed if MoVAE is a good classifier for pure supervised learning tasks.

\paragraph{Datasets.} In this set of experiments and in the next one (Section 4.2) we have considered two datasets: (1) the MNIST dataset\footnote{http://yann.lecun.com/exdb/mnist/. Last visit on 25$^{th}$ September 2017.} of handwritten digits on which the goal is to perform digits recognition, and (2) Fashion-MNIST~\cite{xiao2017FashionMNIST} dataset, released in August 2017 as an alternative to MNIST, on which the goal is to perform fashion products recognition. As MNIST digits dataset is widely known, further on we just briefly describe the Fashion-MNIST dataset~\cite{xiao2017FashionMNIST}. Similarly, with MNIST it has 60000 training images, 10000 testing images - each image being in a gray-scale matrix of size 28x28. The images are  split equally in 10 classes. Differently from MNIST, the images are not digits. Instead they represent fashion products collected from Zalando website and processed to be in the same format as MNIST. The classes of Fashion-MNIST are considerably more difficult than the ones of MNIST as suggested by their names: (0) T-Shirt/Top, (1) Trouser, (2) Pullover, (3) Dress, (4) Coat, (5) Sandals, (6) Shirt, (7) Sneaker, (8) Bag, and (9) Ankle boots. On both datasets, we have used the standard training set for training the models and the standard test set to evaluate them. 

\paragraph{Implementation and evaluation details.} MoVAE has been implemented with the help of the Keras library~\cite{chollet2015}. In all the experiments performed in this set and in the next one (Section~4.2), for a fair and easy comparison, we have used MoVAE models with the same architectures. More exactly, each MoVAE had 10 VAEs, one for each class. Each VAE had 784 input dimensions, 256 intermediate dimensions, and 50 latent dimensions, and dense layers, following the standard VAE example from the Keras library~\cite{chollet2015}. At each iteration, the MoVAE training was done using the RMSProp optimizer~\cite{rmsproptraining} for 40 epochs. The metric used to measure the similarity between an original image and its reconstructed version was Pearson Correlation Coefficient (PCC). In this set of experiments and in the next one (Section 4.2), we have evaluated MoVAE against the default Convolutional Neural Network (CNN) offered as an example in the Keras library\footnote{https://github.com/fchollet/keras/blob/master/examples/mnist\_cnn.py. Last visit on 9$^{th}$ September 2017.}, and state-of-the-art results for that specific task. 

In this set of experiments we used the complete labeled training set of MNIST and Fashion-MNIST to train models. As for this goal the \textit{generalization iterations} phase of MoVAE is not useful (lines 7-21 of Algorithm 1), we set the number of generalization iterations to 0 to skip it. We did not performed any technique to try improving the results, such as data augmentation, and we evaluate it on the standard test sets of both datasets. 

Table~\ref{tab:alltraining} presents the results. We may observe that on both datasets, MoVAE models achieves good accuracies, very close to the ones obtained by the CNN models. In fact, on the MNIST dataset a classification accuracy of 97.85\% is over or, at least, on par with the accuracies obtained by many state-of-the-art machine learning models. The Fashion-MNIST dataset is very new, so no many results are reported yet in the literature, but still MoVAE achieves a classification accuracy which outperforms most of the models reported as benchmark in the launching paper of the dataset~\cite{xiao2017FashionMNIST}. For instance, gradient boosting achieves in average 84\% accuracy. We have good reasons to believe that MoVAE accuracy could be easily improved with fine-tuning, but our goal in this set of experiments was just to prove that MoVAE is a good classifier in a pure supervised learning fashion. This being said, further on we present the most interesting set of experiments, in which we analyze MoVAE performance in the one-shot semi-supervised learning context. 

\begin{table}[b!]
\caption{Purely supervised learning - MoVAE and CNN classification accuracy when the complete labeled training sets of MNIST and Fashion-MNIST were used to train them.}
\tabcolsep=0.06cm
\footnotesize
\begin{tabular}{|c|c|c|}
\hline
Model&MNIST&Fashion-MNIST\\
&Accuracy [\%]&Accuracy [\%]\\
\cline{1-3}
\hline
\hline
MoVAE (ours)&97.9&86.9\\
CNN&\textbf{98.9}&\textbf{92.3}\\
\hline
\end{tabular}
\label{tab:alltraining}
\end{table}

\subsection{One-shot semi-supervised learning}

In this set of experiments, we have evaluated MoVAE in a more difficult setting. We considered just 1, 5, and 10 randomly chosen labeled samples per class for each dataset (from here comes the \textit{one-shot learning} part of the section name). All the other samples belonging to the training sets were used as unlabeled data (from here comes the \textit{semi-supervised learning} part of the section name). We have used the same datasets and implementations as in the previous set of experiments (Section 4.1), with the exception that now we have used also the \textit{generalization iterations} phase of Algorithm~1. The amount of unlabeled data selected by MoVAE at each generalization iteration was set to 3000 samples. 

For each of the three situations (1-shot, 5-shot, and 10-shot), we have repeated the experiments 10 times and we present the averaged results with mean and standard deviation.   

\begin{figure}[t!]  
\centering
\includegraphics[width=1\columnwidth]{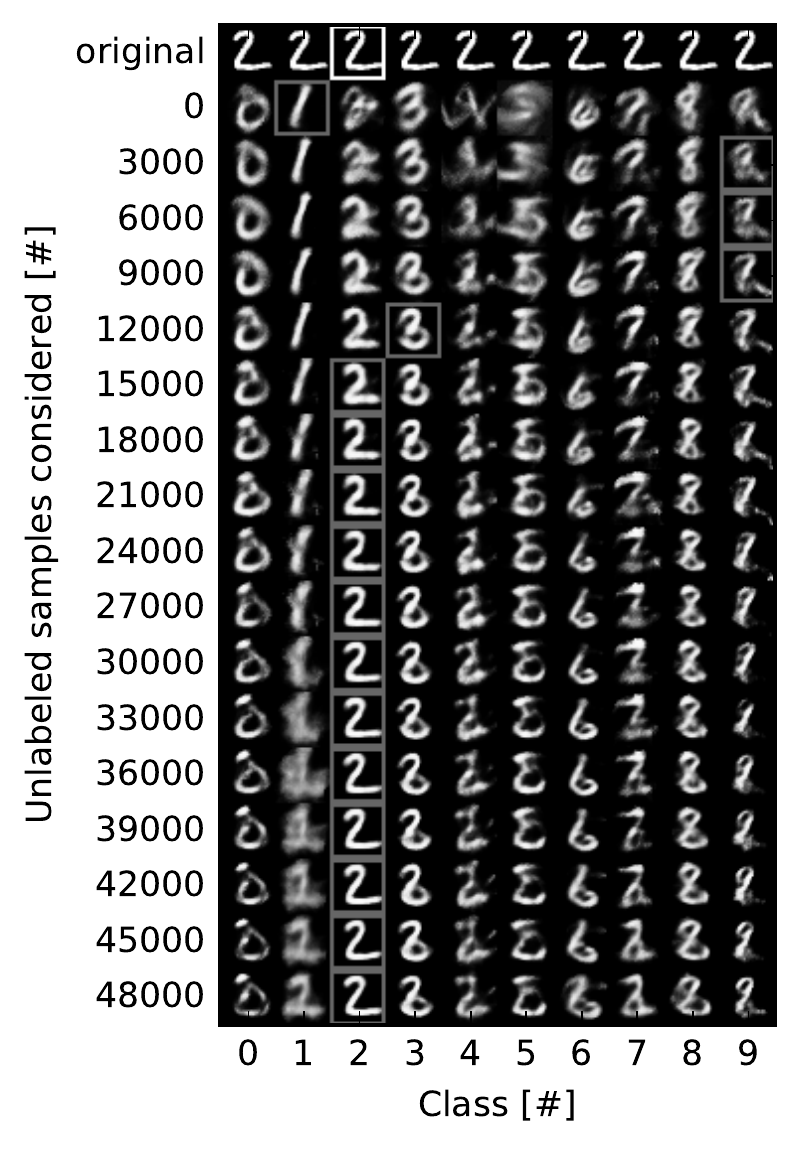}
\caption{One-shot semi-supervised learning - Illustration of MoVAE behavior on the MNIST digits dataset. The first row ("original") shows the original image given to the model, while the following rows show the reconstructed image by each composing VAE (one per class - column) of MoVAE during the generalization iterations. The white square from the first row shows the truth class, while the gray squares from the other rows shows the class predicted by MoVAE at that specific generalization iteration.}
\label{fig:mnistlearnnice}
\end{figure}

\begin{figure*}[t]  
\centering
\includegraphics[width=1\columnwidth]{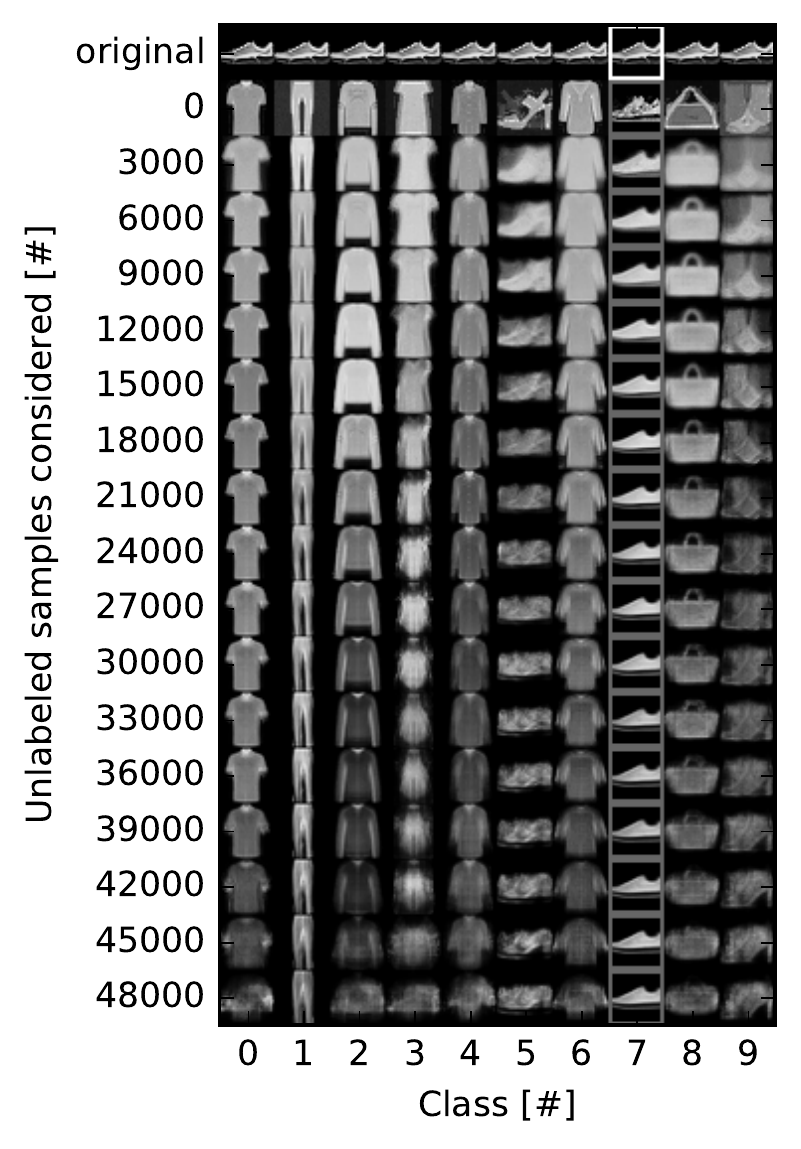}
\includegraphics[width=1\columnwidth]{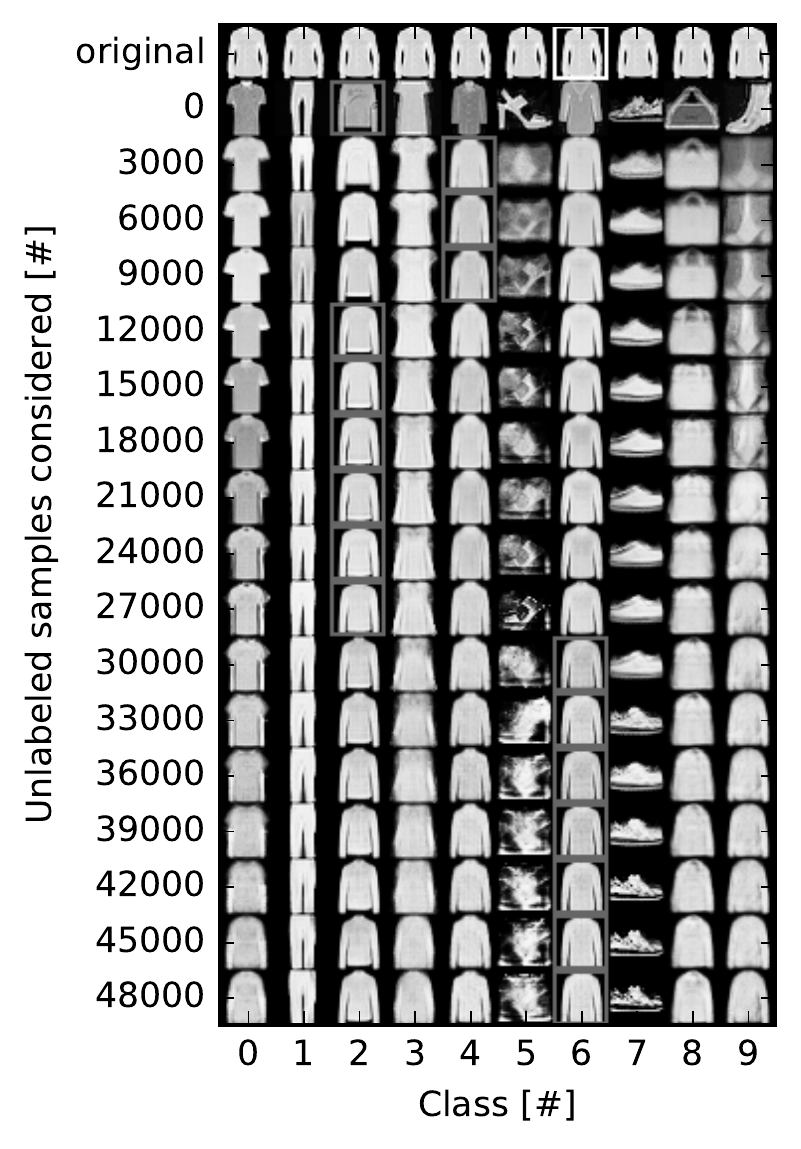}
\vspace{-1.5em}
\caption{One-shot semi-supervised learning - Illustration of MoVAE behavior on the Fashion-MNIST dataset. The first row ("original") shows the original image given to the model, while the following rows show the reconstructed image by each composing VAE (one per class - column) of MoVAE during the generalization iterations. The white square from the first row shows the truth class, while the gray squares from the other rows shows the class predicted by MoVAE at that specific generalization iteration.}
\label{fig:fashionlearnnice}
\end{figure*}

\subsubsection{MNIST digits}

Same as before, first we evaluated MoVAE on the MNIST dataset without using data augmentation techniques. Figure~\ref{fig:mnistoneshot} (left) reflects the evolution of MoVAE accuracy performance and its generalization capabilities while new previously unseen unlabeled samples are taken into consideration. It is interesting to see that in the case when just 1 random chosen labeled sample is considered per class the MoVAE model initially achieves about 45\% mean accuracy. Moreover, during the generalization phase, the MoVAE model is capable to improve itself and to reach about 69\% mean accuracy. Similarly, in the cases where 5 and 10 randomly chosen labeled samples per class are used, MoVAEs show also impressionable generalization capabilities reaching to the end over 90\% mean accuracy. The small standard deviations reflect that MoVAE is a stable model, independently of what labeled samples are given by random chance. In fact, these results would be suffice to overpass state-of-the-art in one-shot learning on the MNIST dataset, as reported in Table~\ref{tab:oneshotmnist}, but further on we show that they can be improved even more, by using data augmentation techniques.

\begin{table}[t!]
\caption{One-shot semi-supervised learning - Classification accuracy of MoVAE against baseline CNN and state-of-the-art using 1, 5, and 10 labeled samples per class on the MNIST and Fashion-MNIST datasets.}
\scriptsize
\tabcolsep=0.06cm
\begin{tabular}{|c|c|c|c|c|c|c|}
\hline
Model&Labeled&Data&Prior&Unlabeled&MNIST&Fashion-MNIST\\
&samples/&Augmen-&Knowledge&Data&Accuracy [\%]&Accuracy [\%]\\
&class [\#]&tation&&&&\\
\hline
\hline
MoVAE (ours)&1-shot&no&no&yes&69.6$\pm$6.5&-\\
MoVAE (ours)&1-shot&yes&no&yes&\textbf{91.1$\pm$4.7}&\textbf{61.6$\pm$2.8}\\
CNN&1-shot&no&no&no&17.4$\pm$3.5&-\\
CNN&1-shot&yes&no&no&22.1$\pm$3.4&21.3$\pm$4.3\\
CPM~\cite{wong2015one}&1-shot&-&yes&no&68.8&-\\
Siamese Net~\cite{Koch2015SiameseNN}&1-shot&-&yes&no&70.3&-\\
Matching Nets~\cite{VinyalsBLKW16}&1-shot&-&yes&no&72.0&-\\
\hline
MoVAE (ours)&5-shot&no&no&yes&90.4$\pm$1.6&-\\
MoVAE (ours)&5-shot&yes&no&yes&\textbf{94.5$\pm$0.6}&\textbf{66.5$\pm$1.7}\\
CNN&5-shot&no&no&no&24.3$\pm$5.4&-\\
CNN&5-shot&yes&no&no&28.1$\pm$5.2&28.2$\pm$4.7\\
CPM~\cite{wong2015one}&5-shot&-&yes&no&83.8&-\\
\hline
MoVAE (ours)&10-shot&no&no&yes&93.1$\pm$1.1&-\\
MoVAE (ours)&10-shot&yes&no&yes&\textbf{94.9$\pm$0.4}&\textbf{70.5$\pm$1.9}\\
CNN&10-shot&no&no&no&33.1$\pm$5.1&-\\
CNN&10-shot&yes&no&no&47.7$\pm$6.6&36.6$\pm$5.4\\
CPM~\cite{wong2015one}&10-shot&-&yes&no&$\approx$88.0&-\\
\hline
\end{tabular}
\label{tab:oneshotmnist}
\end{table}

\paragraph{MNIST digits with data augmentation} More exactly, on the MNIST dataset we have used rotations and small horizontal/vertical shifting. The reason of using data augmentation lays in the wish of creating an initial larger pool of samples (i.e. 500 samples) starting from the 1, 5, or 10 labeled samples considered. Figure~\ref{fig:mnistoneshot} (right) shows that this is a good approach, as MoVAE with just 1 labeled sample per class is capable to achieve an amazing performance of about 91\% mean accuracy on the MNIST dataset, while state-of-the-art best results, up to our best knowledge, report about 72\% accuracy for this scenario. When 5 or 10 random chosen labeled samples are used per class the overall performance increase up to 95\% mean accuracy, this being already very close to the results obtained when the full training set is used as labeled data by MoVAE or other classification models. It worths to be highlighted that with data augmentation the standard deviations become even smaller showing an increasing stability of the MoVAE models.

To understand better MoVAE behavior during learning, in Figure~\ref{fig:mnistlearnnice} we illustrate how each of its composing VAEs is capable to reconstruct a given digit during the generalization learning phase. It may be observed that initially the VAE corresponding to digit 2 of MoVAE is not capable to reconstruct well the given image (i.e. a 2) and in consequence MoVAE fails to classify the given image correctly. Hence, after MoVAE considers 15000 unlabeled samples, it starts generalizing quite well and its corresponding VAE for digit 2 reconstructs the given image well, yielding a correct classification. It is interesting to observe that in this case the other VAEs belonging to the model tend to transform the 2 in their corresponding digit.   

\subsubsection{Fashion-MNIST}

Knowing that data augmentation is a good technique for MoVAE, further on we have used it also on the Fashion-MNIST dataset. More exactly, we have used horizontal flips and a very small zooming. Up to now, there are no results in the literature on the Fashion-MNIST dataset for one-shot learning. Still, our comparisons with the baseline CNN from Table~\ref{tab:oneshotmnist} show that MoVAE is a good model for one-shot learning. To understand better the behavior of our proposed model on this dataset, in Figure~\ref{fig:fashionlearnnice} we give two illustrative examples for MoVAE which uses just 1 labeled sample per class: the reconstruction of a sneaker (left) and the reconstruction of a shirt (right). In the sneaker case the situation is quite simple (all other classes having quite a different shape) and MoVAE is capable to classify it correctly from start without considering any unlabeled data. In the shirt case, the model is bouncing between classifying the given image as a shirt, as a coat, or as a pullover (these three classes being quite similar), but after seeing 30000 unlabeled samples - illustrating the power of the unlabeled data and generalization learning - MoVAE classifies correctly the shirt.

\subsection{One-shot learning}
In the last set of experiments, we have addressed a pure one-shot learning problem. Herein, we did not consider at all unlabeled data, thus we set the number of generalization iterations to 0, ignoring the lines 7-21 of the Algorithm~1. 

\paragraph{Dataset.} We evaluated MoVAE on the Omniglot dataset~\cite{salutomniglot}, a widely used dataset for assessing novel one-shot learning algorithms. Omniglot has in total 50 different alphabets, totalizing 1623 characters. Each character being hand drawn by 20 different people. We mention that the original images have 105x105 pixels and in our experiments we rescale them to 28x28 pixels. We have considered two scenarios, a 5-way and a 1623-way classification problem, with 1 (1-shot) and 5 (5-shot) labeled samples per class (character). The labeled samples from each class were chosen randomly and they were used as training data, while the other samples belonging to same class (19 and 15 samples, respectively) were used as testing data.

In all the experiments performed, we have used data augmentation. By performing a light grid search, we have arrived at the following settings using the Keras data augmentation engine: random rotations with 20 degrees, horizontal and vertical shifts of 0.2, shear range of 0.2, and a zoom range between 0.8 and 1.2. Starting from the labeled samples that we had and with the previous settings we have generated 10000 augmented images for each class. We used these augmented images in the training process. 

\paragraph{Implementation details.} The MoVAE had 5 VAEs for the first scenario, and 1623 VAEs for the second scenario. Each VAE had 784 input dimensions, 784 intermediate dimensions, and 100 latent dimensions, and dense layers. The learning was performed using the RMSProp optimizer~\cite{rmsproptraining} for 50 epochs. As in the previous set of experiments, we observed that the CNN model does not perform well with little labeled data, in this set of experiments we have considered a usual baseline method for one-shot learning, i.e. k-Nearest Neighbours (kNN) with 3 neighbours. We trained and tested MoVAE and kNN on exactly the same data.

\subsubsection{5-way Omniglot}

Usually, one-shot learning on Omniglot is performed in a 5-way manner. This means that just 5 classes, picked at random from the total available number of classes (the ones that are not used to create prior knowledge), are considered to perform the models evaluation~\cite{VinyalsBLKW16}. This setting leads to a random guess accuracy of 20\%. We repeated the experiments 10 times, and we report the results with mean and standard deviation in Table~\ref{tab:oneshotomniglot}. The results show that MoVAE achieves a very good performance, i.e. over 90\% in the case of 1-shot learning, and over 96\% in the case of 5-shot learning. In some of the runs MoVAE achieved even 100\% accuracy. Moreover, the small standard deviation of MoVAE accuracy in comparison with the one of kNN shows that MoVAE is a very stable method. We highlight that MoVAE obtained an accuracy on par with the state-of-the-art Siamese Net~\cite{Koch2015SiameseNN} and Matching Nets~\cite{VinyalsBLKW16} methods, while having the advantage of not needed large amounts of labeled data to build prior knowledge.

\begin{table}[t!]
\caption{One-shot learning - The classification performance of MoVAE and state-of-the-art one-shot learning models on the 5-way Omniglot.}
\footnotesize
\tabcolsep=0.06cm
\begin{tabular}{|c|c|c|c|c|c|}
\hline
Model&Labeled&Data&Prior&Unlabeled&5-way\\
&samples/&Augmen-&Knowledge&Data&Omniglot\\
&class [\#]&tation&&&Accuracy [\%]\\
\hline
\hline
MoVAE (ours)&1-shot&yes&no&no&90.9$\pm$5.4\\
kNN&1-shot&yes&no&no&72.4$\pm$11.6\\
Siamese Net~\cite{Koch2015SiameseNN}&1-shot&yes&yes&no&96.7\\
Matching Nets~\cite{VinyalsBLKW16}&1-shot&yes&yes&no&\textbf{98.1}\\
\hline
MoVAE (ours)&5-shot&yes&no&no&96.7$\pm$2.8\\
kNN&5-shot&yes&no&no&85.2$\pm$8.1\\
Siamese Net~\cite{Koch2015SiameseNN}&5-shot&yes&yes&no&98.4\\
Matching Nets~\cite{VinyalsBLKW16}&5-shot&yes&yes&no&\textbf{98.9}\\
\hline
\end{tabular}
\label{tab:oneshotomniglot}
\vspace*{-1em}
\end{table}

\subsubsection{1623-way Omniglot}

One may argue that MoVAE is not a scalable method, as it needs to build a VAE for each class. To show that this is not the case, in our final experiments we have evaluated it by performing an 1623-way one-shot classification on Omniglot. Indeed, this means that we have built 1623 VAEs, one for each class (character). We highlight that there are no results reported in the literature for 1623-way one-shot classification on Omniglot and we propose this problem as a more realistic benchmark for one-shot learning. 

Siamese Net~\cite{Koch2015SiameseNN} and Matching Nets~\cite{VinyalsBLKW16} methods are not capable of performing 1623-way classification on Omniglot as they need all the data from some classes to create the prior knowledge. An exception for this situation would be the use of other datasets to create the prior knowledge. However, to the best of our knowledge, there are no results reported in the literature for this situation. We mention that this is a clear advantage of MoVAE, as it can perform directly 1623-way classification on Omniglot, and we compare its performance against kNN in Table~\ref{tab:oneshotomniglot1623}. For 1-shot learning MoVAE achieved about 27\% accuracy. At a first thought this does not mean too much, but having in mind that the random guess accuracy is just 0.06\%, this means that MoVAE accuracy is 463 times higher than the one of random guess. Also, it is 9 times higher than the one of kNN. As in the previous experiments, more labeled data means better results, and in the case of 5-shot classification MoVAE performance rises up to 43\%.

In terms of computational time, MoVAE was very fast given the problem considered (i.e. 1623-way Omniglot classification). The total training time for one run was about 4 hours using a standard computer and a M40 GPU. In fact, MoVAE running time was several times faster than the one of kNN.

\begin{table}[t!]
\caption{One-shot learning - MoVAE and kNN classification performance on the 1623-way Omniglot (a new challenge for one-shot learning).}
\footnotesize
\tabcolsep=0.06cm
\begin{tabular}{|c|c|c|c|c|c|}
\hline
Model&Labeled&Data&Prior&Unlabeled&1623-way\\
&samples/&Augmen-&Knowledge&Data&Omniglot\\
&class [\#]&tation&&&Accuracy [\%]\\
\hline
\hline
MoVAE (ours)&1-shot&yes&no&no&\textbf{27.8$\pm$0.4}\\
kNN&1-shot&yes&no&no&3.1$\pm$0.01\\
Random guess&-&-&-&-&0.06\\
\hline
MoVAE (ours)&5-shot&yes&no&no&\textbf{43.2$\pm$0.1}\\
kNN&5-shot&yes&no&no&5.9$\pm$0.01\\
Random guess&-&-&-&-&0.06\\

\hline
\end{tabular}
\label{tab:oneshotomniglot1623}
\vspace*{-1em}
\end{table}

\subsection{Discussion}

To the end, we report that besides MoVAE, we have tried to build a model with similar principles for a different generative model type, i.e. Restricted Boltzmann Machines (RBMs)~\cite{originalrbm}. We have tested the mixture model based on RBMs using just the full labeled training set of MNIST and we obtained a mean accuracy of about 95\%. This accuracy being lower than the one obtained with MoVAE, we have decided to make the thorough analyze just for MoVAE. Auxiliary, in terms of measuring the distance between an original image and its reconstructed version we have tested also the Root Mean Squared Error (RMSE). However, the accuracy results obtained with RMSE were always smaller with 1-3\% than the ones obtained with PCC.

\section{Conclusion}

In this paper, we introduce MoVAE (Mixture of Variational Autoencoders), taking inspiration from the human world. MoVAE is capable to successfully perform the one-shot learning task, without the need of having prior knowledge, due to its generalization learning capabilities. In terms of performance, it goes much beyond the state-of-the-art one-shot learning methods, being capable to overpass them with more than 25\% in accuracy on the MNIST digits recognition task. Besides that, MoVAE achieves very good results (over 60\% accuracy) on a much more complicated real-world task, fashion products recognition, using just 1 labeled sample per class. 

Even when unlabeled data is unavailable, MoVAE is capable of good performance. Thus, we introduce a new challenge for one-shot learning, 1623-way 1-shot learning classification on Omniglot, on which MoVAE accuracy is 463 times higher than the one of random guess. Also, it is 9 times higher than the one of kNN, while no other state-of-the-art results are reported in this very difficult context. This good behavior opens MoVAE path to real-world applications. Overall, our results hint at a promising avenue of research in the attempt of bringing closer the learning capabilities of machine learning models to the ones of humans in the case of collaborative learning just from few examples. 

There are a number of possible future developments following the ideas introduced in this paper. Further on, we enumerate just a few, considering also some limitations of MoVAE: (1) Even MoVAE shows to be very successfully in one-shot learning without prior knowledge, we believe that by making use of prior knowledge, its performance can be further improved; (2) Due to MoVAE generalization capabilities, we believe that if we would combine it with generative replay~\cite{generativereplay2016}, it may be adapted to perform on-line learning; (3) It may be that by creating a VAE for each class will generate on some high dimensional datasets too many connections to optimize, and thus we intend to look into techniques to decrease the amount of these connections for MoVAE before training, such as in~\cite{Mocanu2016xbm}; and (4) We see benefits for plenty real-world applications.



\begin{thebibliography}{00}


\ifx \showCODEN    \undefined \def \showCODEN     #1{\unskip}     \fi
\ifx \showDOI      \undefined \def \showDOI       #1{#1}\fi
\ifx \showISBNx    \undefined \def \showISBNx     #1{\unskip}     \fi
\ifx \showISBNxiii \undefined \def \showISBNxiii  #1{\unskip}     \fi
\ifx \showISSN     \undefined \def \showISSN      #1{\unskip}     \fi
\ifx \showLCCN     \undefined \def \showLCCN      #1{\unskip}     \fi
\ifx \shownote     \undefined \def \shownote      #1{#1}          \fi
\ifx \showarticletitle \undefined \def \showarticletitle #1{#1}   \fi
\ifx \showURL      \undefined \def \showURL       {\relax}        \fi
\providecommand\bibfield[2]{#2}
\providecommand\bibinfo[2]{#2}
\providecommand\natexlab[1]{#1}
\providecommand\showeprint[2][]{arXiv:#2}

\bibitem[\protect\citeauthoryear{Agrawal, Carreira, and Malik}{Agrawal
  et~al\mbox{.}}{2015}]%
        {7410370}
\bibfield{author}{\bibinfo{person}{P. Agrawal}, \bibinfo{person}{J. Carreira},
  {and} \bibinfo{person}{J. Malik}.} \bibinfo{year}{2015}\natexlab{}.
\newblock \showarticletitle{Learning to See by Moving}. In
  \bibinfo{booktitle}{{\em 2015 IEEE International Conference on Computer
  Vision (ICCV)}}. \bibinfo{pages}{37--45}.
\newblock
\showDOI{%
\url{https://doi.org/10.1109/ICCV.2015.13}}


\bibitem[\protect\citeauthoryear{Chollet}{Chollet}{2015}]%
        {chollet2015}
\bibfield{author}{\bibinfo{person}{Fran\c{c}ois Chollet}.}
  \bibinfo{year}{2015}\natexlab{}.
\newblock \bibinfo{title}{keras}.
\newblock \bibinfo{howpublished}{\url{https://github.com/fchollet/keras}}.
  (\bibinfo{year}{2015}).
\newblock


\bibitem[\protect\citeauthoryear{Dalal and Triggs}{Dalal and Triggs}{2005}]%
        {Dalal2005}
\bibfield{author}{\bibinfo{person}{N. Dalal} {and} \bibinfo{person}{B.
  Triggs}.} \bibinfo{year}{2005}\natexlab{}.
\newblock \showarticletitle{Histograms of oriented gradients for human
  detection}. In \bibinfo{booktitle}{{\em 2005 IEEE Computer Society Conference
  on Computer Vision and Pattern Recognition (CVPR'05)}},
  Vol.~\bibinfo{volume}{1}. \bibinfo{pages}{886--893 vol. 1}.
\newblock
\showISSN{1063-6919}
\showDOI{%
\url{https://doi.org/10.1109/CVPR.2005.177}}


\bibitem[\protect\citeauthoryear{Dayan, Hinton, Neal, and Zemel}{Dayan
  et~al\mbox{.}}{1995}]%
        {helmholtzDayan95}
\bibfield{author}{\bibinfo{person}{Peter Dayan}, \bibinfo{person}{Geoffrey~E.
  Hinton}, \bibinfo{person}{Radford~M. Neal}, {and} \bibinfo{person}{Richard~S.
  Zemel}.} \bibinfo{year}{1995}\natexlab{}.
\newblock \bibinfo{title}{The Helmholtz Machine}.
\newblock   (\bibinfo{year}{1995}).
\newblock


\bibitem[\protect\citeauthoryear{Duchi, Hazan, and Singer}{Duchi
  et~al\mbox{.}}{2011}]%
        {rmsproptraining}
\bibfield{author}{\bibinfo{person}{John Duchi}, \bibinfo{person}{Elad Hazan},
  {and} \bibinfo{person}{Yoram Singer}.} \bibinfo{year}{2011}\natexlab{}.
\newblock \showarticletitle{Adaptive Subgradient Methods for Online Learning
  and Stochastic Optimization}.
\newblock \bibinfo{journal}{{\em J. Mach. Learn. Res.\/}}  \bibinfo{volume}{12}
  (\bibinfo{date}{July} \bibinfo{year}{2011}), \bibinfo{pages}{2121--2159}.
\newblock
\showISSN{1532-4435}
\showURL{%
\url{http://dl.acm.org/citation.cfm?id=1953048.2021068}}


\bibitem[\protect\citeauthoryear{Fei-Fei, Fergus, and Perona}{Fei-Fei
  et~al\mbox{.}}{2006}]%
        {FeiFei2006}
\bibfield{author}{\bibinfo{person}{Li Fei-Fei}, \bibinfo{person}{R. Fergus},
  {and} \bibinfo{person}{P. Perona}.} \bibinfo{year}{2006}\natexlab{}.
\newblock \showarticletitle{One-shot learning of object categories}.
\newblock \bibinfo{journal}{{\em IEEE Transactions on Pattern Analysis and
  Machine Intelligence\/}} \bibinfo{volume}{28}, \bibinfo{number}{4}
  (\bibinfo{year}{2006}), \bibinfo{pages}{594--611}.
\newblock


\bibitem[\protect\citeauthoryear{Gluck, Mercado, and Myers}{Gluck
  et~al\mbox{.}}{2011}]%
        {Generalizationhumans}
\bibfield{author}{\bibinfo{person}{Mark~A. Gluck}, \bibinfo{person}{Eduardo
  Mercado}, {and} \bibinfo{person}{Catherine~E. Myers}.}
  \bibinfo{year}{2011}\natexlab{}.
\newblock \bibinfo{booktitle}{{\em Learning and Memory: From Brain to
  Behavior\/} (\bibinfo{edition}{2nd} ed.)}.
\newblock \bibinfo{publisher}{New York: Worth Publishers}.
\newblock


\bibitem[\protect\citeauthoryear{Goyal, Hu, Liang, Wang, and Xing}{Goyal
  et~al\mbox{.}}{2017}]%
        {corr/Goyal17}
\bibfield{author}{\bibinfo{person}{Prasoon Goyal}, \bibinfo{person}{Zhiting
  Hu}, \bibinfo{person}{Xiaodan Liang}, \bibinfo{person}{Chenyu Wang}, {and}
  \bibinfo{person}{Eric~P. Xing}.} \bibinfo{year}{2017}\natexlab{}.
\newblock \showarticletitle{Nonparametric Variational Auto-encoders for
  Hierarchical Representation Learning}.
\newblock \bibinfo{journal}{{\em CoRR\/}}  \bibinfo{volume}{abs/1703.07027}
  (\bibinfo{year}{2017}).
\newblock


\bibitem[\protect\citeauthoryear{Harari}{Harari}{2015}]%
        {humankind}
\bibfield{author}{\bibinfo{person}{Yuval~Noah Harari}.}
  \bibinfo{year}{2015}\natexlab{}.
\newblock \bibinfo{booktitle}{{\em Sapiens: A Brief History of Humankind}}.
\newblock


\bibitem[\protect\citeauthoryear{Hariharan and Girshick}{Hariharan and
  Girshick}{2016}]%
        {HariharanG16}
\bibfield{author}{\bibinfo{person}{Bharath Hariharan} {and}
  \bibinfo{person}{Ross~B. Girshick}.} \bibinfo{year}{2016}\natexlab{}.
\newblock \showarticletitle{Low-shot visual object recognition}.
\newblock \bibinfo{journal}{{\em CoRR\/}}  \bibinfo{volume}{abs/1606.02819}
  (\bibinfo{year}{2016}).
\newblock
\showURL{%
\url{http://arxiv.org/abs/1606.02819}}


\bibitem[\protect\citeauthoryear{Hinton, Dayan, Frey, and Neal}{Hinton
  et~al\mbox{.}}{1995}]%
        {wakesleepHinton1995}
\bibfield{author}{\bibinfo{person}{Geoffrey~E. Hinton}, \bibinfo{person}{Peter
  Dayan}, \bibinfo{person}{Brendan~J. Frey}, {and} \bibinfo{person}{Radford~M.
  Neal}.} \bibinfo{year}{1995}\natexlab{}.
\newblock \showarticletitle{The wake-sleep algorithm for unsupervised neural
  networks}.
\newblock \bibinfo{journal}{{\em Science\/}}  \bibinfo{volume}{268}
  (\bibinfo{year}{1995}), \bibinfo{pages}{1158--1161}.
\newblock


\bibitem[\protect\citeauthoryear{Kingma and Welling}{Kingma and
  Welling}{2013}]%
        {Kingma2013}
\bibfield{author}{\bibinfo{person}{D.~P. Kingma} {and} \bibinfo{person}{M.
  Welling}.} \bibinfo{year}{2013}\natexlab{}.
\newblock \showarticletitle{Auto-encoding variational Bayes}.
\newblock \bibinfo{journal}{{\em CoRR\/}}  \bibinfo{volume}{arXiv:1312.6114}
  (\bibinfo{year}{2013}).
\newblock


\bibitem[\protect\citeauthoryear{Koch, Zemel, and Salakhutdinov}{Koch
  et~al\mbox{.}}{2015}]%
        {Koch2015SiameseNN}
\bibfield{author}{\bibinfo{person}{Gregory Koch}, \bibinfo{person}{Richard
  Zemel}, {and} \bibinfo{person}{Ruslan Salakhutdinov}.}
  \bibinfo{year}{2015}\natexlab{}.
\newblock \showarticletitle{Siamese Neural Networks for One-shot Image
  Recognition}.
\newblock


\bibitem[\protect\citeauthoryear{Kullback}{Kullback}{1959}]%
        {Kullback59}
\bibfield{author}{\bibinfo{person}{Solomon Kullback}.}
  \bibinfo{year}{1959}\natexlab{}.
\newblock \bibinfo{booktitle}{{\em Information Theory and Statistics}}.
\newblock \bibinfo{publisher}{Wiley}.
\newblock


\bibitem[\protect\citeauthoryear{Lake, Salakhutdinov, and Tenenbaum}{Lake
  et~al\mbox{.}}{2015a}]%
        {salutomniglot}
\bibfield{author}{\bibinfo{person}{Brenden~M. Lake}, \bibinfo{person}{Ruslan
  Salakhutdinov}, {and} \bibinfo{person}{Joshua~B. Tenenbaum}.}
  \bibinfo{year}{2015}\natexlab{a}.
\newblock \showarticletitle{{Human-level concept learning through probabilistic
  program induction}}.
\newblock \bibinfo{journal}{{\em Science\/}} \bibinfo{volume}{350},
  \bibinfo{number}{6266} (\bibinfo{date}{11 Dec.} \bibinfo{year}{2015}),
  \bibinfo{pages}{1332--1338}.
\newblock
\showISSN{1095-9203}
\showDOI{%
\url{https://doi.org/10.1126/science.aab3050}}


\bibitem[\protect\citeauthoryear{Lake, Salakhutdinov, and Tenenbaum}{Lake
  et~al\mbox{.}}{2015b}]%
        {LakeScience2015}
\bibfield{author}{\bibinfo{person}{Brenden~M. Lake}, \bibinfo{person}{Ruslan
  Salakhutdinov}, {and} \bibinfo{person}{Joshua~B. Tenenbaum}.}
  \bibinfo{year}{2015}\natexlab{b}.
\newblock \showarticletitle{Human-level concept learning through probabilistic
  program induction}.
\newblock \bibinfo{journal}{{\em Science\/}}  \bibinfo{volume}{350}
  (\bibinfo{date}{12/11/2015} \bibinfo{year}{2015}),
  \bibinfo{pages}{1332--1338}.
\newblock
\showDOI{%
\url{https://doi.org/10.1126/science.aab3050}}


\bibitem[\protect\citeauthoryear{Litjens, Kooi, Bejnordi, Setio, Ciompi,
  Ghafoorian, van~der Laak, van Ginneken, and S{\'{a}}nchez}{Litjens
  et~al\mbox{.}}{2017}]%
        {corr/LitjensKBSCGLGS17}
\bibfield{author}{\bibinfo{person}{Geert J.~S. Litjens}, \bibinfo{person}{Thijs
  Kooi}, \bibinfo{person}{Babak~Ehteshami Bejnordi}, \bibinfo{person}{Arnaud
  Arindra~Adiyoso Setio}, \bibinfo{person}{Francesco Ciompi},
  \bibinfo{person}{Mohsen Ghafoorian}, \bibinfo{person}{Jeroen A. W.~M. van~der
  Laak}, \bibinfo{person}{Bram van Ginneken}, {and} \bibinfo{person}{Clara~I.
  S{\'{a}}nchez}.} \bibinfo{year}{2017}\natexlab{}.
\newblock \showarticletitle{A Survey on Deep Learning in Medical Image
  Analysis}.
\newblock \bibinfo{journal}{{\em CoRR\/}}  \bibinfo{volume}{abs/1702.05747}
  (\bibinfo{year}{2017}).
\newblock
\showURL{%
\url{http://arxiv.org/abs/1702.05747}}


\bibitem[\protect\citeauthoryear{Loncomilla, del Solar, and
  Martínez}{Loncomilla et~al\mbox{.}}{2016}]%
        {LONCOMILLA2016499}
\bibfield{author}{\bibinfo{person}{Patricio Loncomilla},
  \bibinfo{person}{Javier~Ruiz del Solar}, {and} \bibinfo{person}{Luz
  Martínez}.} \bibinfo{year}{2016}\natexlab{}.
\newblock \showarticletitle{Object recognition using local invariant features
  for robotic applications: A survey}.
\newblock \bibinfo{journal}{{\em Pattern Recognition\/}}  \bibinfo{volume}{60}
  (\bibinfo{year}{2016}), \bibinfo{pages}{499 -- 514}.
\newblock
\showISSN{0031-3203}
\showDOI{%
\url{https://doi.org/10.1016/j.patcog.2016.05.021}}


\bibitem[\protect\citeauthoryear{Miller, Matsakis, and Viola}{Miller
  et~al\mbox{.}}{2000}]%
        {MillerCVPR2000}
\bibfield{author}{\bibinfo{person}{Erik~G. Miller},
  \bibinfo{person}{Nicholas~E. Matsakis}, {and} \bibinfo{person}{Paul~A.
  Viola}.} \bibinfo{year}{2000}\natexlab{}.
\newblock \showarticletitle{Learning from One Example through Shared Densities
  on Transforms}. In \bibinfo{booktitle}{{\em 2000 Conference on Computer
  Vision and Pattern Recognition {(CVPR} 2000), 13-15 June 2000, Hilton Head,
  SC, {USA}}}. \bibinfo{pages}{1464--1471}.
\newblock


\bibitem[\protect\citeauthoryear{Mocanu, Mocanu, Nguyen, Gibescu, and
  Liotta}{Mocanu et~al\mbox{.}}{2016a}]%
        {Mocanu2016xbm}
\bibfield{author}{\bibinfo{person}{Decebal~Constantin Mocanu},
  \bibinfo{person}{Elena Mocanu}, \bibinfo{person}{Phuong~H. Nguyen},
  \bibinfo{person}{Madeleine Gibescu}, {and} \bibinfo{person}{Antonio Liotta}.}
  \bibinfo{year}{2016}\natexlab{a}.
\newblock \showarticletitle{A topological insight into restricted Boltzmann
  machines}.
\newblock \bibinfo{journal}{{\em Machine Learning\/}} \bibinfo{volume}{104},
  \bibinfo{number}{2} (\bibinfo{date}{01 Sep} \bibinfo{year}{2016}),
  \bibinfo{pages}{243--270}.
\newblock
\showISSN{1573-0565}
\showDOI{%
\url{https://doi.org/10.1007/s10994-016-5570-z}}


\bibitem[\protect\citeauthoryear{Mocanu, Vega, Eaton, Stone, and Liotta}{Mocanu
  et~al\mbox{.}}{2016b}]%
        {generativereplay2016}
\bibfield{author}{\bibinfo{person}{Decebal~Constantin Mocanu},
  \bibinfo{person}{Maria~Torres Vega}, \bibinfo{person}{Eric Eaton},
  \bibinfo{person}{Peter Stone}, {and} \bibinfo{person}{Antonio Liotta}.}
  \bibinfo{year}{2016}\natexlab{b}.
\newblock \showarticletitle{Online Contrastive Divergence with Generative
  Replay: Experience Replay without Storing Data}.
\newblock \bibinfo{journal}{{\em CoRR\/}}  \bibinfo{volume}{abs/1610.05555}
  (\bibinfo{year}{2016}).
\newblock


\bibitem[\protect\citeauthoryear{Ouyang, Wang, Zeng, Qiu, Luo, Tian, Li, Yang,
  Wang, Loy, and Tang}{Ouyang et~al\mbox{.}}{2015}]%
        {7298854}
\bibfield{author}{\bibinfo{person}{W. Ouyang}, \bibinfo{person}{X. Wang},
  \bibinfo{person}{X. Zeng}, \bibinfo{person}{Shi Qiu}, \bibinfo{person}{P.
  Luo}, \bibinfo{person}{Y. Tian}, \bibinfo{person}{H. Li},
  \bibinfo{person}{Shuo Yang}, \bibinfo{person}{Zhe Wang},
  \bibinfo{person}{Chen-Change Loy}, {and} \bibinfo{person}{X. Tang}.}
  \bibinfo{year}{2015}\natexlab{}.
\newblock \showarticletitle{DeepID-Net: Deformable deep convolutional neural
  networks for object detection}. In \bibinfo{booktitle}{{\em 2015 IEEE
  Conference on Computer Vision and Pattern Recognition (CVPR)}}.
  \bibinfo{pages}{2403--2412}.
\newblock
\showISSN{1063-6919}
\showDOI{%
\url{https://doi.org/10.1109/CVPR.2015.7298854}}


\bibitem[\protect\citeauthoryear{Rezende, Mohamed, and Wierstra}{Rezende
  et~al\mbox{.}}{2014}]%
        {pmlr-v32-rezende14}
\bibfield{author}{\bibinfo{person}{Danilo~Jimenez Rezende},
  \bibinfo{person}{Shakir Mohamed}, {and} \bibinfo{person}{Daan Wierstra}.}
  \bibinfo{year}{2014}\natexlab{}.
\newblock \showarticletitle{Stochastic Backpropagation and Approximate
  Inference in Deep Generative Models}. In \bibinfo{booktitle}{{\em Proceedings
  of the 31st International Conference on Machine Learning}} {\em
  (\bibinfo{series}{Proceedings of Machine Learning Research})},
  \bibfield{editor}{\bibinfo{person}{Eric~P. Xing} {and} \bibinfo{person}{Tony
  Jebara}} (Eds.), Vol.~\bibinfo{volume}{32}. \bibinfo{publisher}{PMLR},
  \bibinfo{address}{Bejing, China}, \bibinfo{pages}{1278--1286}.
\newblock
\showURL{%
\url{http://proceedings.mlr.press/v32/rezende14.html}}


\bibitem[\protect\citeauthoryear{Salimans, Kingma, and Welling}{Salimans
  et~al\mbox{.}}{2015}]%
        {icml2015_salimans15}
\bibfield{author}{\bibinfo{person}{Tim Salimans}, \bibinfo{person}{Diederik
  Kingma}, {and} \bibinfo{person}{Max Welling}.}
  \bibinfo{year}{2015}\natexlab{}.
\newblock \showarticletitle{Markov Chain Monte Carlo and Variational Inference:
  Bridging the Gap}. In \bibinfo{booktitle}{{\em Proceedings of the 32nd
  International Conference on Machine Learning (ICML-15)}}.
  \bibinfo{publisher}{JMLR Workshop and Conference Proceedings},
  \bibinfo{pages}{1218--1226}.
\newblock


\bibitem[\protect\citeauthoryear{Santoro, Bartunov, Botvinick, Wierstra, and
  Lillicrap}{Santoro et~al\mbox{.}}{2016}]%
        {SantoroBBWL16}
\bibfield{author}{\bibinfo{person}{Adam Santoro}, \bibinfo{person}{Sergey
  Bartunov}, \bibinfo{person}{Matthew Botvinick}, \bibinfo{person}{Daan
  Wierstra}, {and} \bibinfo{person}{Timothy~P. Lillicrap}.}
  \bibinfo{year}{2016}\natexlab{}.
\newblock \showarticletitle{One-shot Learning with Memory-Augmented Neural
  Networks}.
\newblock \bibinfo{journal}{{\em CoRR\/}}  \bibinfo{volume}{abs/1605.06065}
  (\bibinfo{year}{2016}).
\newblock
\showURL{%
\url{http://arxiv.org/abs/1605.06065}}


\bibitem[\protect\citeauthoryear{Smolensky}{Smolensky}{1987}]%
        {originalrbm}
\bibfield{author}{\bibinfo{person}{P. Smolensky}.}
  \bibinfo{year}{1987}\natexlab{}.
\newblock \showarticletitle{Information Processing in Dynamical Systems:
  Foundations of Harmony Theory}.
\newblock  (\bibinfo{year}{1987}), \bibinfo{pages}{194--281}.
\newblock


\bibitem[\protect\citeauthoryear{Thewlis, Bilen, and Vedaldi}{Thewlis
  et~al\mbox{.}}{2017}]%
        {thewlis17Bunsupervised}
\bibfield{author}{\bibinfo{person}{J. Thewlis}, \bibinfo{person}{H. Bilen},
  {and} \bibinfo{person}{A. Vedaldi}.} \bibinfo{year}{2017}\natexlab{}.
\newblock \showarticletitle{Unsupervised object learning from dense invariant
  image labelling}. In \bibinfo{booktitle}{{\em Proceedings of Advances in
  Neural Information Processing Systems (NIPS)}}.
\newblock


\bibitem[\protect\citeauthoryear{Vinyals, Blundell, Lillicrap, Kavukcuoglu, and
  Wierstra}{Vinyals et~al\mbox{.}}{2016}]%
        {VinyalsBLKW16}
\bibfield{author}{\bibinfo{person}{Oriol Vinyals}, \bibinfo{person}{Charles
  Blundell}, \bibinfo{person}{Timothy~P. Lillicrap}, \bibinfo{person}{Koray
  Kavukcuoglu}, {and} \bibinfo{person}{Daan Wierstra}.}
  \bibinfo{year}{2016}\natexlab{}.
\newblock \showarticletitle{Matching Networks for One Shot Learning}.
\newblock  (\bibinfo{year}{2016}), \bibinfo{pages}{3630--3638}.
\newblock


\bibitem[\protect\citeauthoryear{Waterman}{Waterman}{1970}]%
        {WATERMAN1970}
\bibfield{author}{\bibinfo{person}{D.A. Waterman}.}
  \bibinfo{year}{1970}\natexlab{}.
\newblock \showarticletitle{Generalization learning techniques for automating
  the learning of heuristics}.
\newblock \bibinfo{journal}{{\em Artificial Intelligence\/}}
  \bibinfo{volume}{1}, \bibinfo{number}{1} (\bibinfo{year}{1970}),
  \bibinfo{pages}{121 -- 170}.
\newblock
\showISSN{0004-3702}


\bibitem[\protect\citeauthoryear{Wong and Yuille}{Wong and Yuille}{2015}]%
        {wong2015one}
\bibfield{author}{\bibinfo{person}{Alex Wong} {and} \bibinfo{person}{Alan~L
  Yuille}.} \bibinfo{year}{2015}\natexlab{}.
\newblock \showarticletitle{One shot learning via compositions of meaningful
  patches}. In \bibinfo{booktitle}{{\em Proceedings of the IEEE International
  Conference on Computer Vision}}. \bibinfo{pages}{1197--1205}.
\newblock


\bibitem[\protect\citeauthoryear{Woodward and Finn}{Woodward and Finn}{2017}]%
        {WoodwardF17}
\bibfield{author}{\bibinfo{person}{Mark Woodward} {and}
  \bibinfo{person}{Chelsea Finn}.} \bibinfo{year}{2017}\natexlab{}.
\newblock \showarticletitle{Active One-shot Learning}.
\newblock \bibinfo{journal}{{\em CoRR\/}}  \bibinfo{volume}{abs/1702.06559}
  (\bibinfo{year}{2017}).
\newblock
\showURL{%
\url{http://arxiv.org/abs/1702.06559}}


\bibitem[\protect\citeauthoryear{Xiao, Rasul, and Vollgraf}{Xiao
  et~al\mbox{.}}{2017}]%
        {xiao2017FashionMNIST}
\bibfield{author}{\bibinfo{person}{Han Xiao}, \bibinfo{person}{Kashif Rasul},
  {and} \bibinfo{person}{Roland Vollgraf}.} \bibinfo{year}{2017}\natexlab{}.
\newblock \bibinfo{title}{Fashion-MNIST: a Novel Image Dataset for Benchmarking
  Machine Learning Algorithms}.
\newblock   (\bibinfo{year}{2017}).
\newblock


\end{thebibliography}


\end{document}